# A Propound Method for the Improvement of Cluster Quality

Shveta Kundra Bhatia[1], V.S. Dixit[2]

[1] Computer Science Department, Swami Shraddhanand College, University of Delhi,
New Delhi 110036, India
*shvetakundra@gmail.com*

[2] Computer Science Department, Atma Ram Sanatam Dharam College, University of Delhi,
New Delhi 110010, India
*veersaindixit@rediffmail.com*

**Abstract**

In this paper Knockout Refinement Algorithm (KRA) is proposed to refine original clusters obtained by applying SOM and K-Means clustering algorithms. KRA Algorithm is based on Contingency Table concepts. Metrics are computed for the Original and Refined Clusters. Quality of Original and Refined Clusters are compared in terms of metrics. The proposed algorithm (KRA) is tested in the educational domain and results show that it generates better quality clusters in terms of improved metric values.

## 1. Introduction

Clustering or grouping of users or sessions into logically meaningful clusters is a well studied branch of research. Clustering has been formulated in various ways by using algorithms such as K-Means, SOM etc. K-Means partitions the objects into clusters by minimizing the sum of squared distances between the objects and the centroids of the clusters. Many algorithms have been proposed to improve implementation of the K-Means algorithm.

1.1 Literature Review

Bentley (1975) suggested kd-trees to improve triangle inequality to enhance K-Means. Bradley and Fayyad (1998) presented a procedure for computing a refined starting condition that is based on a technique for estimating the modes of a distribution to converge to a better local minimum that can be coupled with scalable clustering algorithms. A Kernel-Means algorithm was established by Scholkops in 1998 that maps data points from the input space to a feature space through a non linear transformation minimizing the clustering error in feature space. Samet (1999) devised R-Trees which were not appropriate for problems with high dimensions. A partial distance algorithm (PDA) had been proposed in 2000 which allows termination of the distance calculation at an early stage for better results. In 2002 Dhillon suggested refining clusters in high dimensional text data by combining the first variation principal and spherical K-Means. Elkan in 2003 suggested the use of triangle inequality to accelerate K-Means. Refinement of clusters from K-Means with Ant Colony Optimization was suggested by Mary, Raja in 2005. In 2010 refinement of web usage data clustering from K-Means with Genetic Algorithm was suggested by N. Sujatha and K. Iyakutty and in 2011 by Prabha, Saranya. Various initialization methods such as Binary Splitting was reviewed together with several other initialization techniques and compared with KKZ. Multilevel refinement schemes were used for refining and improving the clusters produced by hierarchical agglomerative clustering by Karypis. Self Organizing Feature Maps (SOM) is a type of Artificial Neural Network that is trained using Unsupervised Learning which helps in visualizing high dimensional data into low dimensional data. Researchers have contributed to the improvement of various SOM quality measures such as quantization error, topographic product, topographic error, trustworthiness and neighborhood preservation. The researchers have contributed only to accelerate the algorithms; there is not much contribution in refinement of clusters. In this paper we try to refine the original clusters generated by SOM and K-Means algorithms. We proposed a Knockout Refinement Algorithm (KRA) to refine original clusters using a contingency table to compute dissimilarity among sessions in clusters and eliminating sessions with a higher count of dissimilarity. We then calculate the Davies Bouldin (DB) index, Dunn's Index, Precision, Recall and F-Measure for the original and refined clusters.

1.2 Standard K-Means Algorithm

K-means works using the following steps:
1. Place K objects points into the space that are to be clustered by choosing the initial value of K. Object points represent centroids of initial groups chosen.
2. Assign each object point to the group that has *the closest Centroids.*
3. Re-compute the positions of the K centroids and continue till all object points have been assigned.
4. Repeat Steps 2 and 3 until the centroids do not change. This produces a separation of the object points into groups from which the metric to be minimized can be calculated.

The algorithm aims to minimize an objective function as in Eq. (1)

$$J = \sum_{j=1} \sum_{i=1} ||x_i^{(j)} - c_j||^2 \qquad (1)$$

Where, $\left\|x_i^{(j)} - c_j\right\|^2$ is a chosen distance measure between a data point and the cluster centre. It indicates the distance of the n data points from their respective cluster centers.

1.3 Standard SOM Algorithm

1. Assign random values to the weight vectors of a neuron.
2. Provide an input vector to the network.
3. Traverse each node in the network
   a) Find similarity between the input vector and the network's node's weight vector using Euclidean Distance.
   b) Find the node that produces the smallest distance which is assigned as the Best Matching Unit (BMU).
4. Update the nodes in the neighborhood of the BMU by changing the weights using Eq. (2):

$$Wv(t + 1) = Wv(t) + \Theta(t)\alpha(t)(D(t) - Wv(t)) \qquad (2)$$

Where,
- t keeps an account of the iteration number
- λ is the iteration range
- **Wv** is the current weight vector
- D is the target input
- Θ(*t*) is the Gaussian neighborhood function
- α(*t*) is learning rate due to time

5. Increment t and repeat from step 2 while *t*< λ.

1.4 Metrics

1.4.1 Davies Bouldin Index

This index aims to identify sets of clusters that are compact and well separated. The Davies-Bouldin index is defined as in Eq. (3):

$$DB = 1/K \sum_{i=1}^{K} \max\left[\frac{diam(C_i) + diam(C_j)}{d(C_i, C_j)}\right] \qquad (3)$$

Were K denotes the number of clusters, $i, j$ are cluster labels, $diam(C_i)$ and $diam(C_j)$ are the diameters of the clusters $C_i$ and $C_j$, $d(C_i, C_j)$ is the average distance between the clusters. Smaller values of average similarity between each cluster and its most similar one indicate a "better" clustering solution.

1.4.2 Dunn's Index

This index aims at expecting a large distance between the clusters and an expected small diameter of the cluster. The index is defined as in Eq. (4):

$$D_x = \min_{i=1\ldots x}\{\min_{j=1\ldots x}\{d(c_i, c_j) | \max_{k=1\ldots x} diam(c_k)\}|\} \qquad (4)$$

Here $d(c_i, c_j)$ is to compute the dissimilarity between two clusters defined as in Eq. (5):

$$d(c_i, c_j) = \min_{a \in c_i, b \in c_j} d(a, b) \qquad (5)$$

And diam(c) is the diameter of the cluster that defines the maximum distance between two points in a cluster. A large value of Dunn's index indicates compact and well separated clusters.

1.4.3 F-Measure, Precision And Recall

F-measure combines the precision and recall concepts from information retrieval. We then calculate the recall and precision of that cluster for each class as:

$$Recall(i, j) = x_{ij} / x_i$$

And
$$Precision(i, j) = x_{ij} / x_j$$

Were $x_{ij}$ is the number of objects of class $i$ that are in cluster $j$, $x_j$ is the number of objects in cluster $j$, and $x_i$, is the number of objects in class $i$. Precision and Recall are measures that help to evaluate the quality of a set of retrieved documents. The $F - Measure$ of cluster $j$ and class $i$ is given by the following Eq. (6):

F (*i*, *j*) = 2 *Recall(*i*, *j*)Precision(*i*, *j*) / Precision(*i*, *j*) + Recall(*i*, *j*) (6)

The $F - Measure$ values are within the interval [0, 1] and larger values indicate higher clustering quality.

## 2. Process Description and Experiments

unique URLs. 72.9% of the total sessions were exported into a .csv format with the help of scripts in tcl language as rest of the sessions had only either one or two page views.

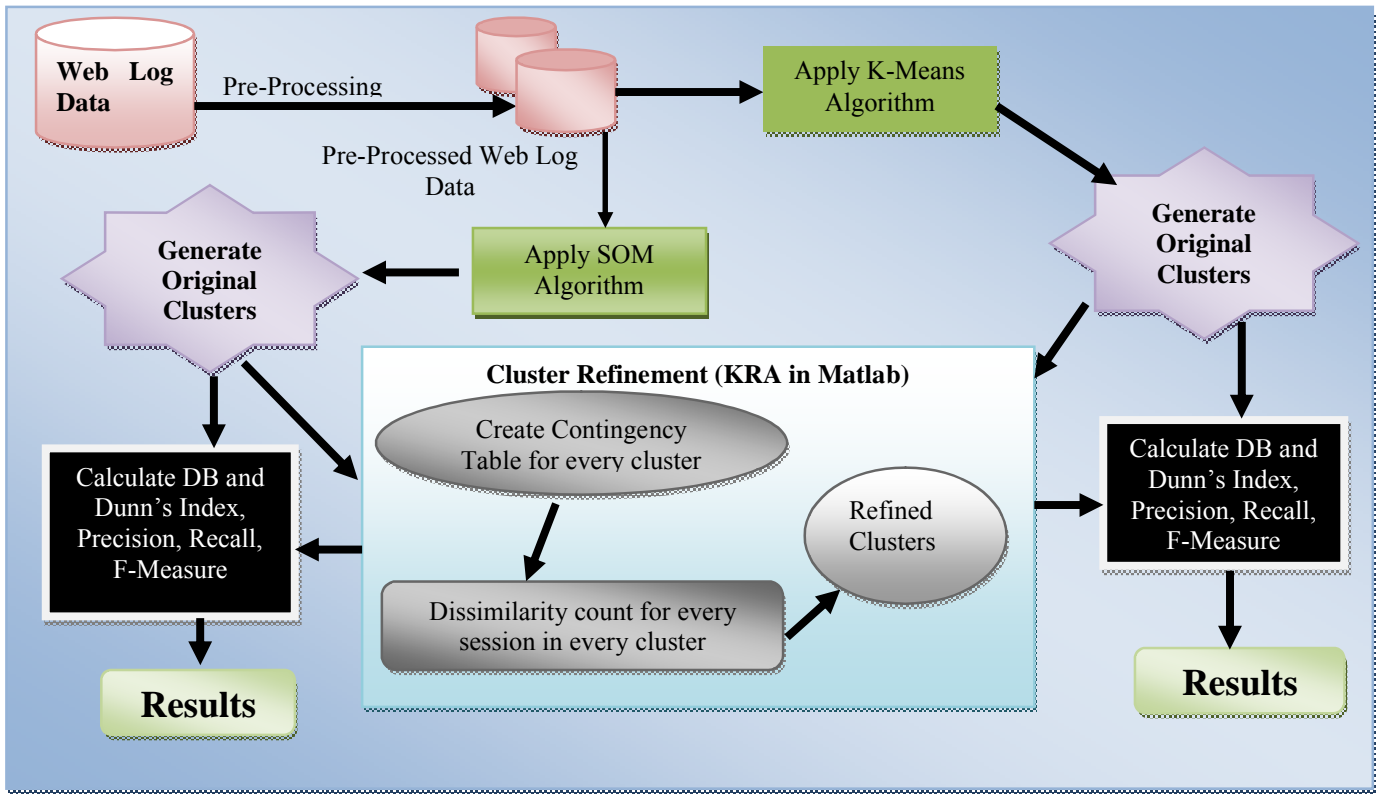

Fig 1: Architecture of proposed refinement algorithm

The raw web log file we used for the experiment contained 5999 web requests that can be found at http://www.vtsns.edu.rs/maja/vtsnsNov16 containing information like date/time of request, hit type, page, hostname, referrer, server domain, authenticated user, server response, page views, size etc. For preparing the web log data for the mining process, it needs to be cleared of irrelevant requests; and transformed to a format that can be fed into the clustering algorithm. For Pre-Processing and creation of sessions a tool called Sawmill (Version 7.0), developed by Flower fire Inc. has been used. Sawmill computes session information by tracking the page, date/time, and visitor id for each page view. When a session view is requested, it processes all of these page views at the time of the request. Sawmill groups the hits into initial sessions based on the visitor id by assuming that each visitor contributes to a session. In the next step sorting by time is performed for a click-by-click record of each visitor. A session timeout interval of 30 minutes is considered for generating final sessions and sessions longer than 2 hours are eliminated. Using the Sawmill tool on our web log data led to the creation of sessions and 110

Further we optimized our matrix and 59.1% of the sessions and 43 unique URLs were used for experimentation. The optimization was performed on the basis of sessions having less than 3 page views and pages that were viewed 5 or less than 5 times have been removed. The optimized matrix was used for clustering using the Self-Organizing Feature Maps and K-Means algorithms. We used the Spice-SOM tool and SPSS software for implementation of the respective algorithms. Applying the two algorithms we can see that clusters with similarity among sessions have been obtained and can be used for prediction of pages to a user of similar interests. Clusters of sizes 10, 15 and 20 were generated using both the techniques of K-Means and SOM. Apply KRA to refine the original clusters. The quality of obtained clusters is evaluated using the Davies Bouldin and Dunn's quality measure along with external quality measures such as Precision, Recall and F-Measure. The clusters are listed as Original Clusters (OC) on which we shall apply our proposed Knockout Refinement Algorithm (KRA).

# 3. Proposed Knockout Refinement Algorithm (KRA)

**Input:** Set of Original Clusters (OC).

**Process: Step 1:** Do
Pick up the first cluster ($C_k$)
Generate contingency table for every pair of sessions C ($S_i$, $S_j$) in the cluster.
Evaluate dissimilarity between all sessions using the following equation:
$$d_k(s_i,s_j) = \sum_{\substack{k=1 \\ i \neq j}}^{n} r_{ki} + s_{ki}/q_{ki} + r_{ki} + s_{ki}$$

Generate a Symmetric Dissimilarity Matrix(SDM)
    If Threshold > 0.3 for d(Si,Sj)
      Count++
        If(Count>2)
          Eliminate Si and Sj from the cluster $C_k$
        End If
    End If
    Generate Refined Clusters (RC)
End Do
Repeat for all the clusters

**Output**: Refined Clusters (RC)

Fig 2: Proposed KRA Algorithm

Where, Dissimilarity can be computed using contingency table which is a 2 by 2 matrix between two sessions.

|  | Session j | | |
|---|---|---|---|
|  | 1 | 0 | Sum |
| Session i   1 | q | r | q + r |
| 0 | s | t | s + t |
| Sum | q + s | r + t |  |

Where q is the number of variables that equal 1 for both i and j sessions, r is the number of variables that equal 1 for session i but that are equal to 0 for session j, s is the number of variables that equal 0 for session i and equal 1 for session j and t is the number of variables that equal 0 for sessions i and j. In this case the factor t is unimportant and is ignored to compute the dissimilarity. The dissimilarity for all sessions is computed as in equation 7:

$$d_k(s_i,s_j) = \sum_{\substack{k=1 \\ i \neq j}}^{n} r_{ki} + s_{ki}/q_{ki} + r_{ki} + s_{ki} \quad (7)$$

The above computation is applied to every pair of sessions that generates a Symmetric Dissimilarity Matrix (SDM) as follows:

|  | S1 | S2 | S3 | S4 |
|---|---|---|---|---|
| S1 | 0 | d(S1, S2) | d(S1,S3) | d(S1,S4) |
| S2 | d(S2, S1) | 0 | d(S2, S3) | d(S2,S4) |
| S3 | d(S3, S1) | d(S3,S2) | 0 | d(S3,S4) |
| S4 | d(S4, S1) | d(S4, S2) | d(S4, S3) | 0 |

For refinement using the above Symmetric Dissimilarity Matrix (SDM) and using a threshold value of 0.3 session pairs are counted and those sessions are removed from the cluster whose count values are greater than 2. Performing the above computation we get refined clusters.

# 4. Outcome

## 4.1 Results for Davies Bouldin Index

The results for comparison of DB index for original and refined clusters are as follows:

Table 1: Comparison of DB Index for K-Means Algorithm

| DAVIES BOULDIN INDEX (K-MEANS) | | |
|---|---|---|
|  | ORIGINAL CLUSTERS | REFINED CLUSTERS |
| 10 Clusters | 1.6868 | 1.6338 |
| 15 Clusters | 1.8329 | 1.6315 |
| 20 Clusters | 1.5459 | 1.5486 |

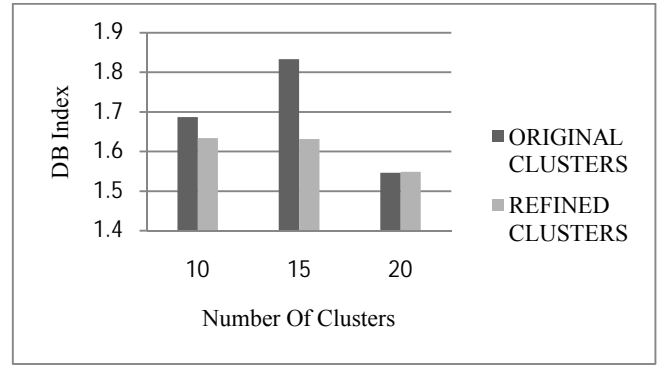

Fig 3: Comparison of DB Index for K-Means Algorithm

Table 2: Comparison of DB index for SOM Algorithm

| DAVIES BOULDIN INDEX (SOM) | | |
|---|---|---|
|  | ORIGINAL CLUSTERS | REFINED CLUSTERS |
| 10 Clusters | 2.5719 | 2.4217 |
| 15 Clusters | 2.4185 | 1.3263 |
| 20 Clusters | 1.9049 | 1.7203 |

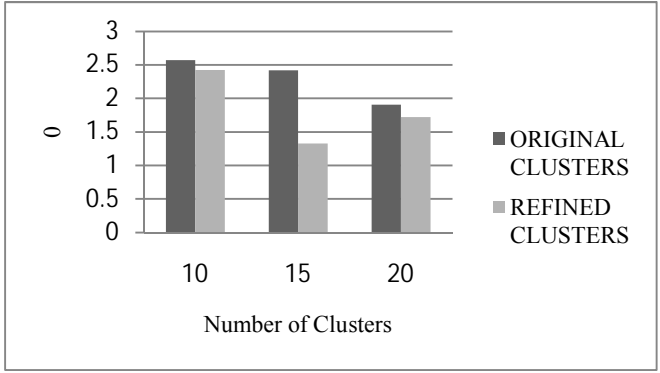

Fig 4: Comparison of DB Index for K-Means Algorithm

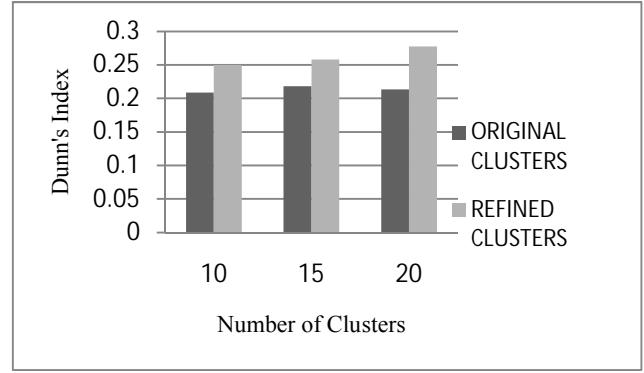

Fig 6: Comparison of Dunn's Index for SOM Algorithm

## 4.2 Results For Dunn's Index

The results for comparison of Dunn's index for original and refined clusters are as follows:

Table 3: Comparison of Dunn's Index for K-Means Algorithm

| DUNN'S INDEX (K-MEANS) | | |
|---|---|---|
| | ORIGINAL CLUSTERS | REFINED CLUSTERS |
| 10 Clusters | 0.2425 | 0.2773 |
| 15 Clusters | 0.2581 | 0.2581 |
| 20 Clusters | 0.2672 | 0.2886 |

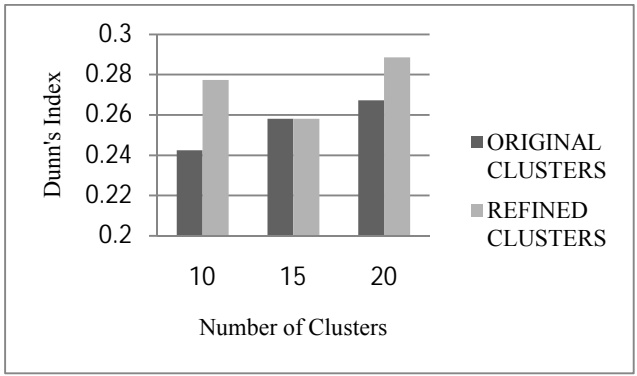

Fig 5: Comparison Of DB Index for K-Means Algorithm

Table 4: Comparison of Dunn's Index for SOM Algorithm

| DUNN'S INDEX (SOM) | | |
|---|---|---|
| | ORIGINAL CLUSTERS | REFINED CLUSTERS |
| 10 Clusters | 0.2085 | 0.25 |
| 15 Clusters | 0.2182 | 0.2581 |
| 20 Clusters | 0.2132 | 0.2773 |

## 4.3 Results For Precision, Recall And F-Measure

Table 5: Comparison of Precision, Recall and F-Measure for 10 Clusters Using K-Means

| K-MEANS(10 Clusters) | | |
|---|---|---|
| | ORIGINAL CLUSTERS | REFINED CLUSTERS |
| Precision | 0.9548 | 1 |
| Recall | 0.2512 | 0.2535 |
| F-Measure | 0.3977 | 0.4045 |

Table 6: Comparison of Precision, Recall and F-Measure for 10 Clusters Using SOM

| SOM(10 Clusters) | | |
|---|---|---|
| | ORIGINAL CLUSTERS | REFINED CLUSTERS |
| Precision | 0.8881 | 0.9131 |
| Recall | 0.0767 | 0.0767 |
| F-Measure | 0.1413 | 0.1416 |

Table 7: Comparison of Precision, Recall and F-Measure for 15 Clusters Using K-Means

| K-MEANS(15 Clusters) | | |
|---|---|---|
| | ORIGINAL CLUSTERS | REFINED CLUSTERS |
| Precision | 0.909 | 0.9201 |
| Recall | 0.2357 | 0.2419 |
| F-Measure | 0.3743 | 0.383 |

Table 8: Comparison Of Precision, Recall And F-Measure For 15 Clusters Using SOM

| SOM(15 Clusters) | | |
|---|---|---|
| | ORIGINAL CLUSTERS | REFINED CLUSTERS |
| Precision | 0.6679 | 0.6846 |
| Recall | 0.0915 | 0.0992 |
| F-Measure | 0.1609 | 0.1733 |

Table 9: Comparison Of Precision, Recall And F-Measure For 20 Clusters Using K-Means

| K-MEANS(20 Clusters) | | |
|---|---|---|
| | ORIGINAL CLUSTERS | REFINED CLUSTERS |
| Precision | 0.9125 | 0.9208 |
| Recall | 0.2116 | 0.2116 |
| F-Measure | 0.3436 | 0.3442 |

Table 10: Comparison Of Precision, Recall And F-Measure For 20 Clusters Using SOM

| SOM(20 Clusters) | | |
|---|---|---|
| | ORIGINAL CLUSTERS | REFINED CLUSTERS |
| Precision | 0.7437 | 0.777 |
| Recall | 0.0919 | 0.093 |
| F-Measure | 0.1635 | 0.1662 |

## 5. CONCLUSION

The proposed algorithm tested on the web log data shows that refined clusters lead to an improved Davies Bouldin and Dunn's Index values; external quality measures such as Precision, Recall and F-Measure have improved for refined clusters as compared to the original clusters. The proposed algorithm is scalable and can be coupled with clustering algorithms to address any other web log data sets. Note that the performances of clustering algorithms are found to be data dependent.

**Shveta Kundra Bhatia** is working as an Assistant Professor in the Department Of Computer Science, Swami Sharaddhanand College, University of Delhi. Her research area is Web Usage Mining and is currently pursuing PhD under Dr. V.S. Dixit from Department of Computer Science, University of Delhi.

**Dr. V. S. Dixit** is working in the Department Of Computer Science, Atma Ram Sanatan Dharam College, University of Delhi. His research area is Queuing theory, Peer to Peer systems, Web Usage Mining and Web Recommender Systems. He is currently engaged in doing the research. He is Life member of IETE.